\documentclass[final]{cvpr}

\usepackage{times}
\usepackage{epsfig}
\usepackage{graphicx}
\usepackage{amsmath}
\usepackage{amssymb}
\usepackage{multirow}
\usepackage{unicode}

\usepackage{adjustbox}

\usepackage{amsfonts}       

\usepackage{times}
\usepackage{epsfig}
\usepackage{graphicx}
\usepackage{amsmath}
\usepackage{amssymb}
\usepackage{multirow}
\usepackage{color}
\usepackage{algorithmic}
\usepackage{algorithm}
\usepackage{booktabs}
\usepackage{url}
\usepackage[table]{xcolor}
\usepackage{colortbl}
\usepackage{bm}
\usepackage{subcaption}
\usepackage{makecell}
\usepackage{pifont}
\usepackage{enumitem}

\usepackage[pagebackref=true,breaklinks=true,colorlinks,bookmarks=false]{hyperref}

\newcommand{\cmark}{\ding{51}}%
\newcommand{\xmark}{\ding{55}}%

\def\cvprPaperID{2839} 
\def\confYear{CVPR 2021}

\begin{document}

\title{Bipartite Graph Network with Adaptive Message Passing \\ 
for Unbiased Scene Graph Generation}

\author{Rongjie Li\textsuperscript{\rm 1,3,4}
	\quad Songyang Zhang \textsuperscript{\rm 1,3,4}
   \quad Bo Wan \textsuperscript{\rm 1,5,} \thanks{This work was done when Bo Wan was a master student in ShanghaiTech University.This work was supported by Shanghai NSF Grant (No. 18ZR1425100). 
   Code is available: \url{https://github.com/Scarecrow0/BGNN-SGG} }
	\quad Xuming He\textsuperscript{\rm 1,2} \\
	\textsuperscript{\rm 1}School of Information Science and Technology, ShanghaiTech University \quad \\
	\textsuperscript{\rm 2}Shanghai Engineering Research Center of Intelligent Vision and Imaging\\
	\textsuperscript{\rm 3}Shanghai Institute of Microsystem and Information Technology,
	Chinese Academy of Sciences\\
	\textsuperscript{\rm 4}University of Chinese Academy of Sciences\\
	\textsuperscript{\rm 5}Department of Electrical Engineering(ESAT), KU Leuven\\
	\{lirj2, zhangsy, wanbo, hexm\}@shanghaitech.edu.cn
}

\maketitle

\begin{abstract}
   Scene graph generation is an important visual understanding task with a broad range of vision applications. Despite recent tremendous progress, it remains challenging due to the intrinsic long-tailed class distribution and large intra-class variation.
   To address these issues, we introduce a novel confidence-aware bipartite graph neural network with adaptive message propagation mechanism for unbiased scene graph generation. 
   In addition, we propose an efficient bi-level data resampling strategy to alleviate the imbalanced data distribution problem in training our graph network. 
   Our approach achieves superior or competitive performance over previous methods on several challenging datasets, including Visual Genome, Open Images V4/V6, demonstrating its effectiveness and generality.


\end{abstract}


\section{Introduction}

\begin{figure}
    \centering
    \includegraphics[width=\linewidth]{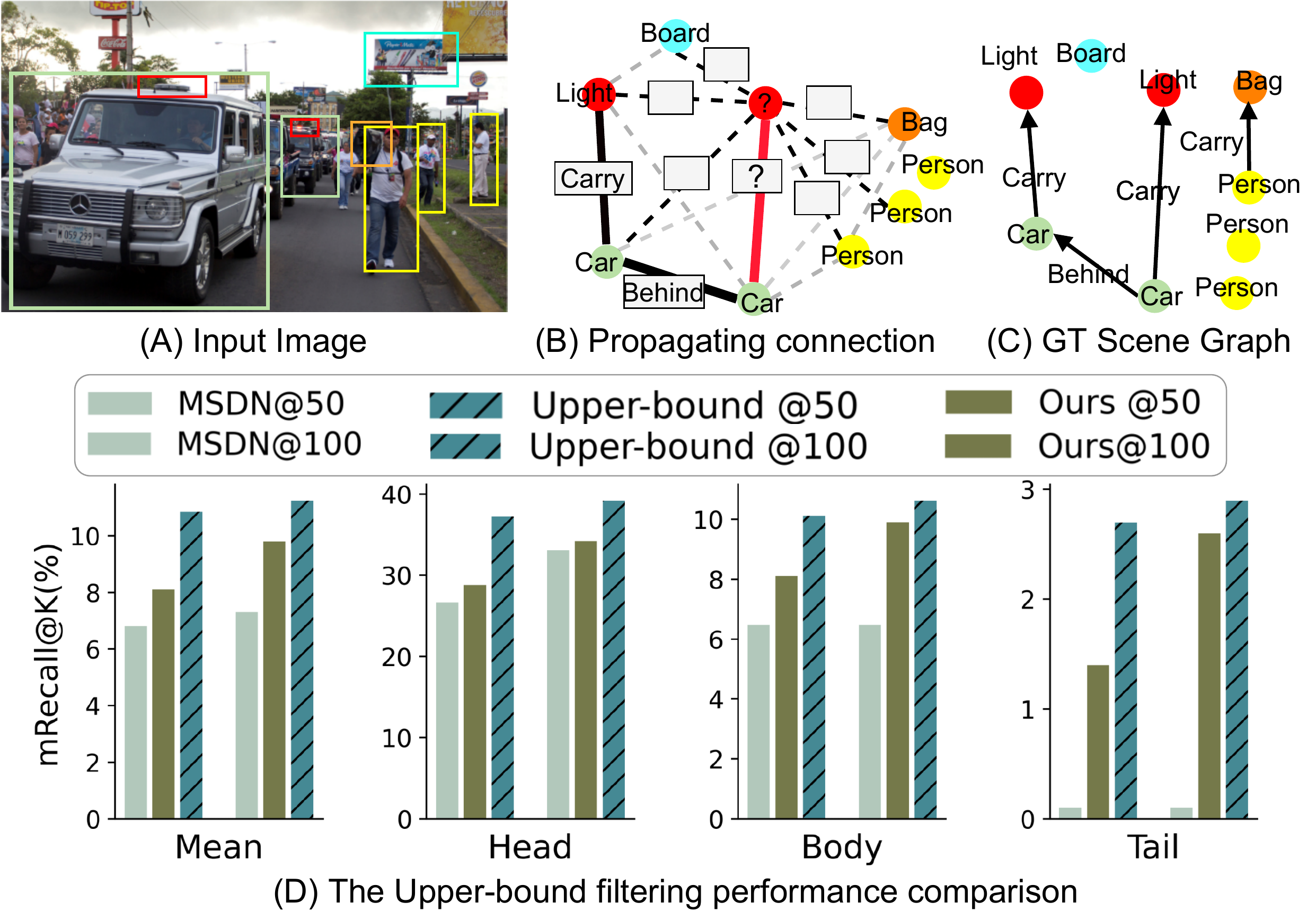}
    \caption{
        \textbf{The illustration of biased scene graph generation and empirical study on Visual Genome.} As shown in (D), the baseline (MSDN~\cite{li_scene_2017}) performance is dominated by the head categories due to the imbalanced data. We estimate an upper-bound performance by ignoring negative predicate-entity connections during message propagation, as shown in (B). Its performance (shown in (D)) indicates a large room for improvement in context modeling.
    }
    \label{fig:ad}
\end{figure}



Scene graph generation, which aims to detect visual objects and their relationships (or \textit{triplets: <subject, predicate, object>}) in an image, is a fundamental visual understanding task. Such a compact structural scene representation has potential applications in many vision tasks such as visual question answering~\cite{teney2017graph, shi2019explainable, hildebrandt2020scene}, image captioning~\cite{yang2019auto} and image retrieval~\cite{johnson2015image}. 
Tremendous progress has been made recently in scene graph generation~\cite{krishna2017visual, xu_scene_2017, li_scene_2017, zellers_neural_2017, yang_graph_2018, li_factorizable_2018, tang_learning_2018,chen_knowledge-embedded_2019, gu2019scene, zhang_graphical_2019, tang_unbiased_2020, lin_gps-net_2020, wang_sketching_2020}, thanks to learned visual representations and advances in object detection.  
However, this task remains particularly challenging due to large variations in visual relationships, extremely imbalanced object and relation distribution and lack of sufficient annotations for many categories.  

One primary challenge, which causes \textit{biased relationship prediction}, is the intrinsic long-tail data distribution. A scene graph model has to simultaneously cope with imbalanced annotations among the head and medium-sized categories, and few-shot learning in the tail categories. A naively learned model will be largely dominated by those few head categories with much degraded performance for many tail categories (as shown in Fig.~\ref{fig:ad}-D). Early work~\cite{cao_learning_2019,cui_class-balanced_2019} on re-balancing data distribution focus on data re-sampling or loss re-weighting. However, it is non-trivial to directly apply the image-level re-balancing strategies for such instance-level tasks. Recent efforts try to introduce the re-balancing ideas into object detection~\cite{gupta_lvis:_2019,tan_equalization_2020} and scene graph generation~\cite{tang_unbiased_2020}, but it remains difficult to achieve a satisfactory trade-off between head and tail categories.



Moreover, those non-head predicate categories typically involve complex semantic meaning and large intra-class variations (\textit{e.g.} play, look) in images, which exacerbates the problems in their representation learning and classification. Many previous works~\cite{zellers_neural_2017, xu_scene_2017, li_scene_2017, tang_learning_2018,yang_graph_2018,lin_gps-net_2020} attempt to address this problem by developing context modeling mechanisms, but often suffer from noisy information propagation due to their use of fully connected graphs. 
More recent efforts~\cite{qi_attentive_2018,yang_graph_2018,tang_learning_2018,wang_sketching_2020} aim to improve context modeling by designing a sparse structure, which also limits the model flexibility. To illustrate the impact of noises in graph, we further conduct an empirical analysis, as shown in Fig.\ref{fig:ad}, which indicates that \textit{a baseline model can achieve notable performance improvement by removing the noisy subject-object associations}.

Based on these findings, we propose a novel confidence-aware graph representation and its learning strategy for unbiased scene graph generation. To this end, we first develop a bipartite graph neural network (BGNN) with the adaptive message propagation for effective context modeling. 
Specifically, our method takes the hypothesize-and-classify strategy, which first generates a set of visual entity and predicate proposals from a proposal generation network. Then we compute a context-aware representation for those proposals by passing them through a multi-stage BGNN. 
Our graph network adopts directed edges to model different information flow between entity and relationship proposals as a bipartite graph, and an adaptive message propagation strategy based on relation confidence estimation to reduce the noise in the context modeling. Finally, we use the refined entity and predicate representations to predict their categories with linear classifiers. 

To train our multi-stage BGNN for unbiased prediction, we also design a bi-level data resampling strategy to alleviate the imbalanced data distribution problem. Our method combines the image-level over-sampling and instance-level under-sampling ideas~\cite{hu_learning_2020,gupta_lvis:_2019} for the structured prediction task. Equipped with this strategy, we can achieve a better trade-off between the head and tail categories and learn our bipartite graph neural network more effectively.



We extensively validate our methods on three scene graph generation datasets, including Visual Genome, Open Images V4, and Open Images V6. 
The empirical results and ablative studies show our method consistently achieves competitive or state-of-the-art performance on all benchmarks. The main contributions of our works are three-folds.
\begin{itemize}[noitemsep,topsep=0pt]
	\item We introduce a bipartite graph neural network with adaptive message propagation to alleviate the error propagation and achieve effective context modeling.
	\item We propose a bi-level data resampling to achieve a better trade-off between head and tail categories for scene graph generation.
    \item Our method achieves competitive or state-of-the-art performance on various scene graph benchmarks.
\end{itemize}

\section{Related Works}

\paragraph{Scene Graph Generation.}
Traditional methods in scene graph generation typically utilize graph-based context-modeling strategies to learn discriminative representation for node and edge prediction. Most of them either focus on the graph structure design or leveraging scene context via various message propagation mechanisms. 

Several types of graph structure have been proposed for context modeling in literature. A popular idea is to model the context based on a sequential model (\textit{e.g.}, LSTM)~\cite{zellers_neural_2017} or a fully-connected graph~\cite{xu_scene_2017,dai_drnet_2017, li_scene_2017,yin_zoom-net:_2018,woo_linknet:_2018, wang_exploring_2019, lin_gps-net_2020}. 
In addition, recent works~\cite{tang_learning_2018, wang2020tackling, yang2019auto, qi_attentive_2018} explore sparse graph structures, which are either associated with the downstream tasks (\textit{e.g.} VQA) or built by trimming the relationship proposals according to the category or geometry information of subject-object pairs. However, these works often rely on their specific designs based on the downstream tasks, which limits the flexibility of their representations. 



Another direction aims to incorporate context information into existing deep ConvNet models by exploring different message propagation mechanisms. A common strategy is to perform message passing between the entities proposals~\cite{zellers_neural_2017, tang_learning_2018, wang_sketching_2020, woo_linknet:_2018, qi_attentive_2018, wang_exploring_2019, lin_gps-net_2020, chen_knowledge-embedded_2019}, while the other aggregates the contextual information between the entities and predicates~\cite{xu_scene_2017, li_scene_2017, dai_drnet_2017, li_factorizable_2018, yin_zoom-net:_2018, yang_graph_2018, wang_exploring_2019, cong_nodis_nodate}, which also produces effective scene graph representations.


Our work considers both message passing and inferring network connectivity in a single framework. In particular, we develop a generic Bipartite Graph Neural Network (BGNN) to effectively model the context of the entity and predicate proposals, and an adaptive message propagation to compute a more flexible representation. The previous SGG models~\cite{li_scene_2017, xu_scene_2017, li_factorizable_2018, yang_graph_2018} can be considered as special cases of the BGNN.

\vspace{-4mm}
\paragraph{Long-tail Visual Recognition}
Previous works in visual recognition typically utilize re-balancing strategies to alleviate biased prediction caused by long-tail distributions.
These re-balancing strategies include dataset resampling to achieve balanced class prior~\cite{chawla_smote_2002, drummond_why_2003, shen_relay_2016, mahajan_exploring_2018}, and loss re-weighting based on instance frequency or hard-example mining~\cite{cao_learning_2019, cui_class-balanced_2019, khan_cost_2017, tan_equalization_2020, lin_focal_2017, ssd_liu_2015, li2020overcoming, lu_gridrcnn_2018}. 
Recently, \cite{hu_learning_2020,gupta_lvis:_2019} propose an instance-level re-sampling strategy for the tasks of object detection and instance segmentation.
Other approaches also explore knowledge transfer learning from head categories for long-tail classification~\cite{liu_largescale_2019, gidaris_dynamic_2018} or develop the two-stage learning scheme~\cite{zhou_bbn_nodate, kang2019decoupling}. 
However, it is non-trivial to apply a naive re-balancing strategy for scene graph generation.
We propose a \textit{bi-level data sampling strategy} by combing the \textit{image-level over-sampling}~\cite{gupta_lvis:_2019} and \textit{instance-level under-sampling}~\cite{hu_learning_2020}  to achieve a better trade-off between the head and tail categories.


For the task of scene graph generation, several strategies have proposed to tackle the intrinsic long-tail problem. Some researchers propose novel loss designs by leveraging the semantic constraints of scene graph~\cite{knyazev_graph_2020, lin_gps-net_2020, yan_pcpl_2020, wang2020tackling}.  Others develop new graph structure encoding the context~\cite{tang_learning_2018, chen_knowledge-embedded_2019, lin_gps-net_2020} or introduce external commonsense and linguistic knowledge~\cite{zareian_bridging_2020, zareian_learning_2020, peyre2019detecting} for better representation learning. Recently, Tang \etal~\cite{tang_unbiased_2020} proposes an unbiased inference method by formulating the recognition process as a causal model. In this work, we aim to improve the context modeling for tail-categories by design an novel graph network and message propagation mechanism.

\begin{figure*}[th]
	\centering
	\includegraphics[width=15cm]{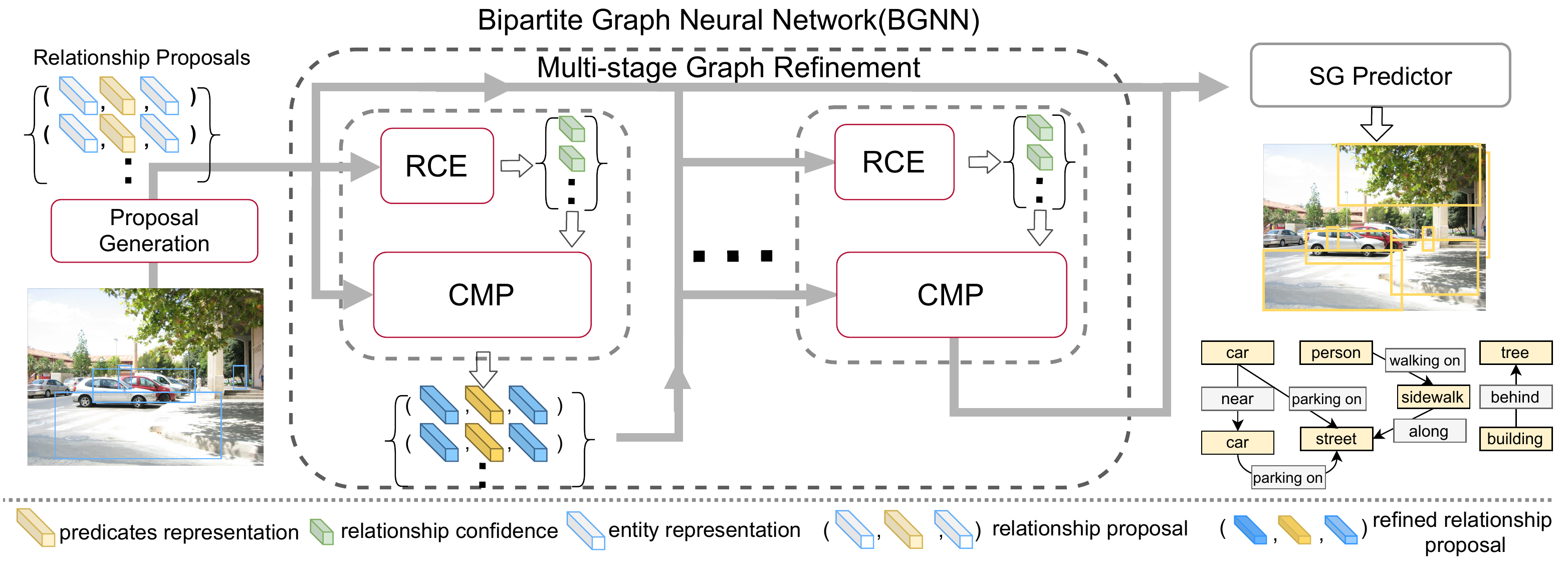}
	\caption{
		\textbf{Illustration of overall pipeline of our BGNN model.}
		\textbf{RCE} denotes the relationship confidence estimation module. 
		\textbf{CMP} denotes the confidence-aware message propagation model.
		\textbf{SG Predictor} is the scene graph predictor for the final prediction.
	}\label{pipeline_overview}
	\vspace{-0.4cm}
\end{figure*}

\vspace{-4mm}
\paragraph{Graph Neural Network}
Our work is also related to learning deep networks on graph-structure data. The Graph Neural Network (GNN) is first proposed by~\cite{scarselli_graph_2009}, which is a powerful method for handling the non-Euclidean data. 
Previous works have explored different graph structures (e.g. directed graph~\cite{kampffmeyer_rethinking_2019}, heterogeneous graph~\cite{zhang_deep_2018, wang_heterogeneous_2019} ) and aggregation mechanism (e.g. convolutional ~\cite{kipf_semi-supervised_2017, hamilton_inductive_2018}, attention~\cite{li_gated_2017, wang_non-local_2017, zhang_latentgnn:_2019, vaswani_attention_2017}) for various tasks. Several recent efforts attempt to improve the quality of message propagation in GNNs.
Hou \etal~\cite{hou_measuring_2020} proposes a context-surrounding GNN framework to measure the quality of neighborhood aggregation.
Xu \etal~\cite{xu_dynamically_2020} introduces a dynamical subgraph construction by applying a graphical attention mechanism conditioned on input queries. 
In this work, we introduce a novel bipartite graph network to learn a robust context-aware feature representation for the predicates.
\section{Our Approach}  


We aim to tackle scene graph generation, which parses an input image into a structural graph representation of object entities and their visual relationship in the scene.  In particular, we focus on addressing the challenge of biased scene graph prediction, mainly caused by the intrinsic long-tail distribution of visual relationship in a typical training dataset and the large intra-class variation of predicate categories. 
To this end, we introduce an adaptive scene graph generation strategy, which simultaneously learns a robust entity/predicate representation and a calibrated classifier for better balanced performance.  

In this section, we first present the problem setting of scene graph generation and an overview of our method in Sec.~\ref{subsec:problem}. We then introduce the details of our predicate representation in Sec.~\ref{subsec:model}, followed by our learning strategy on the predicate classifier and representation in Sec.~\ref{subsec:learning}.



\subsection{Problem Setting and Overview}\label{subsec:problem}

\paragraph{Problem Setting} 
Given an image $\mathbf{I}$, 
the task of scene graph generation (SGG) aims to parse the input $\mathbf{I}$ into a scene graph $\mathcal{G}_{scene}=\{\mathcal{V}_o,\mathcal{E}_r\}$, where $\mathcal{V}_o$ is the node set encoding object entities and $\mathcal{E}_r$ is the edge set that represents predicate between an ordered pair of entities. 
Typically, each node $v_i\in \mathcal{V}_o$ has a category label from a set of entity classes $\mathcal{C}_e$ and a corresponding image location represented by a bounding box, while each edge  
$e_{i\to j} \in \mathcal{E}_r $ between a pair of nodes $v_i$ and $v_j$ is associated with a predicate label from a set of predicate classes $\mathcal{C}_p$ in this task.

%
\vspace{-4mm}
\paragraph{Method Overview}
In this work, we adopt a hypothesize-and-classify strategy for the unbiased scene graph generation. Our approach first generates a set of entity and predicate proposals and then computes their context-aware representations, followed by predicting their categories.

Concretely, we introduce a bipartite graph network that explicitly models the interactions between entities and their predicates in order to cope with unreliable contextual information from the noisy proposals. Based on the graph network, we develop an adaptive message passing scheme capable of actively controlling the information flows to reduce the noise impact in the graph and generate a robust context-aware representation for each relationship proposal.

Taking this representation, we then learn predicate and entity classifiers to predict the categories of predicate and entity within relationship proposals.
To alleviate the bias effect of the imbalanced data, we also design an efficient bi-level data resampling strategy for the model training, which enables us to achieve a better trade-off between the head and tail categories.
An overview of our method is illustrated in Fig.~\ref{pipeline_overview}, and we will start from a detailed description of our model architecture below.

\subsection{Model Architecture}\label{subsec:model}

Our scene graph generation model is a modular deep network consisting of three main submodules: 1) a \textit{proposal generation network} to generate entity and relationship proposals and compute their initial representation (Sec.~\ref{subsec:proposal}); 2) a \textit{bipartite graph neural network} to encode the scene context with adaptive message propagation and multi-stage iterative refinement (Sec.~\ref{subsec:bgnn}); and 3) a \textit{scene graph predictor} to decode the scene graph from the context-aware representations of relationship proposals (Sec.~\ref{subsec:predictor}).

\subsubsection{Proposal Generation Network}
\label{subsec:proposal}

Following~\cite{zellers_neural_2017, zhang_graphical_2019}, we utilize an object detector network (e.g., Faster R-CNN~\cite{ren_faster_2015}) to generate a set of entity and relationship proposals. 
The entity proposals are taken directly from the detection output with their categories and classification scores, while the relationship proposals are generated by forming ordered pairs of all the entity proposals.

Given the relationship proposals, we then compute an initial representation for both entities and predicates. Specifically, for the $i$-th entity proposal, we denote its convolution feature as $\mathbf{v}_i$, its bounding box as $b_i$ and its detected class as $c_i$. The entity representation $\mathbf{e}_i$ uses a fully-connected network $f_e$ to integrate its visual, geometric and semantic features as,   
\begin{align}
    \mathbf{e}_i = f_e(\mathbf{v}_i\oplus \mathbf{g}_i\oplus \mathbf{w}_i)
\end{align}
where $\mathbf{g}_i$ is a geometric feature based on its bounding box $b_i$, $\mathbf{w}_i$ is a semantic feature based on a word embedding of its class $c_i$, and $\oplus$ is the concatenation operation. 

For the relationship proposal from entity $i$ to $j$, we combines the entity representations $\mathbf{e}_i\oplus\mathbf{e}_j$ with the convolutional feature of their union region (denoted as $\mathbf{u}^p_{i,j}$). Formally, we compute the predicate representation $\mathbf{r}_{i\to  j}$ as 
\begin{align}
    \mathbf{r}_{i\to  j}=f_{u}(\mathbf{u}^p_{i,j}) + f_{p}(\mathbf{e}_i\oplus\mathbf{e}_j)
\end{align}
where $f_u$ and $f_p$ are two fully-connected networks.

\subsubsection{Bipartite Graph Neural Network}
\label{subsec:bgnn}

Given the relationship proposals, we build a graph structure to capture the dependency between entities and predicates. To this end, we introduce a bipartite graph $\mathcal{G}_b$ with directed edges, which enables us to model the different information flow directions between entity and predicate representations.
Specifically, the graph consists of two groups of nodes $\mathcal{V}_e,\mathcal{V}_p$, which correspond to entity representations and predicate representations respectively. Those two groups of nodes are connected by two sets of directed edges $\mathcal{E}_{e\to  p}$ and $\mathcal{E}_{p\to  e}$ representing information flows from the entities to predicates and vice versa. Hence the bipartite graph has a form as $\mathcal{G}_b=\{\mathcal{V}_e,\mathcal{V}_p, \mathcal{E}_{e\to  p}, \mathcal{E}_{p\to  e}\}$.

To effectively model the context of the entity and predicate proposals, we develop a Bipartite Graph Neural Network (BGNN) on the graph $\mathcal{G}_b$.   
Our BGNN conducts a multi-stage message propagation and each stage consists of 1) a \textit{relationship confidence estimation} module to provide a confidence estimate on relationship; 2) a \textit{confidence-aware message propagation} to incorporate scene context and semantic cues into the entity/predicate proposals.
The overview of our BGNN is illustrated in Fig.~\ref{pipeline_overview}. We will focus on a single stage of our network in the rest of this section.
 

\vspace{-4mm}
\paragraph{Relationship Confidence Estimation (RCE) Module}\label{subsubsec:rce}
In order to reduce the noise in context modeling, we introduce a relationship confidence estimation (RCE) module. It predicts a confidence score for each relationship proposal to control the information flow in the message propagation.  

Concretely, for a predicate node from entity $i$ to $j$, the RCE module takes as input the predicate proposal features $\mathbf{r}_{i\to  j}$ and its associated entities' class scores, and predicts a confidence score for each predicate class as below, 
\begin{align}
\mathbf{s}^m_{i\to  j}=g_x(\mathbf{r}_{i\to  j}\oplus\mathbf{p}_i\oplus \mathbf{p}_j)\in\mathbb{R}^{|\mathcal{C}_p|}
\end{align}
where $\mathbf{p}_i,\mathbf{p}_j\in \mathbb{R}^{|\mathcal{C}_e|}$ are the class probabilities for entity $\mathbf{e}_i$ and $\mathbf{e}_j$ from the detection, and $ g_x$ is a multilayer fully-connected network. 
We then fuse those confidence scores into a global confidence score for the predicate node as
\begin{align}
s^b_{i\to  j}= \sigma (\mathbf{w}_b^\intercal\mathbf{s}^m_{i\to  j}), \quad \mathbf{w}_b\in\mathbb{R}^{|\mathcal{C}_p|}
\end{align}
where $\sigma$ is the sigmoid activation function and $\mathbf{w}_b$ are the parameters for the fusion.



\begin{figure}
    \centering
    \includegraphics[width=7.6cm]{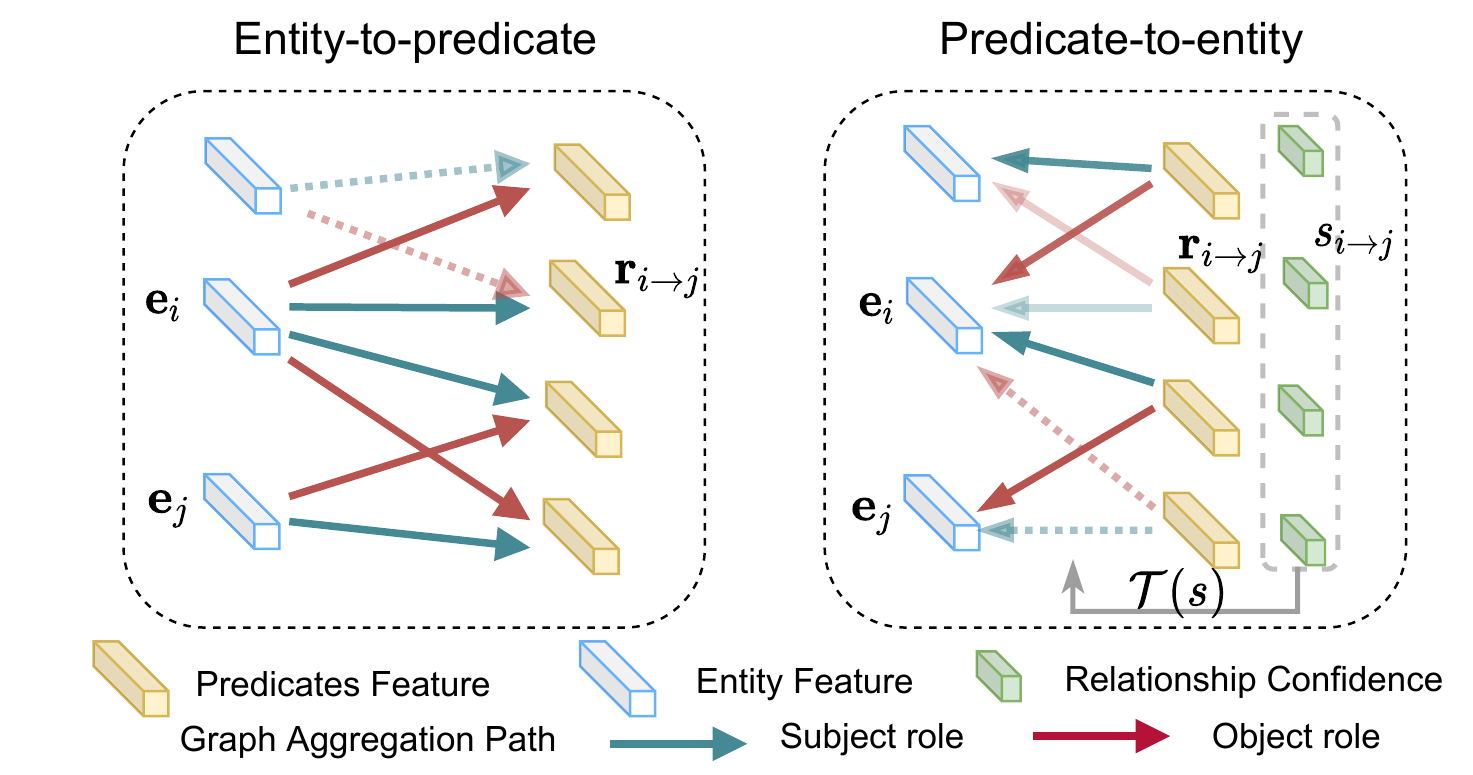}
    \caption{\textbf{Two kinds of message propagation with confidence gating of bipartite graph neural network.}
    The dashed arrows mean information flow is blocked when the source is uncertain.
    The different alpha value stands for different aggregation weights for rest connections.
    }
    \label{fig:confidence_aware_msp} 
    \vspace{-4mm}
\end{figure}

\vspace{-4mm}
\paragraph{Confidence-aware Message Propagation}
We now introduce our adaptive message propagation for capturing the scene context. Specifically, we design two types of message passing update, including an \textit{entity-to-predicate} message and a \textit{predict-to-entity} message according to the edge directions, as illustrated in Fig.~\ref{fig:confidence_aware_msp}. Below we consider an iteration from $l$ to $l+1$ in the message passing.

\paragraph{}{1) Entity-to-predicate Message Propgation:}
We update the representation of a predicate node $\mathbf{r}_{i\to  j}$ by fusing its neighboring entity nodes:
\begin{align}
    &\mathbf{r}_{i\to  j}^{(l+1)} = \mathbf{r}_{i\to  j}^{(l)} + 
        \phi (
                d_{s} \mathbf{W}_{r}^\intercal \mathbf{e}_{i}^{(l)}+d_{o} \mathbf{W}_{r}^\intercal \mathbf{e}_{j}^{(l)}
        )\\
       &d_{s}= \sigma(\mathbf{w}_{s}^\intercal[\mathbf{r}_{i\to  j}^{(l)}\oplus\mathbf{e}_i^{(l)}]), \;
       d_{o}= \sigma(\mathbf{w}_{o}^\intercal[\mathbf{r}_{i\to  j}^{(l)}\oplus\mathbf{e}_j^{(l)}])
\end{align}
where $\mathbf{W}_r$ is a linear transformation, $\phi$ is an activation function (\textit{e.g.} ReLU).\, and $d_s, d_o$ are learnable affinity functions of entity and predicate, where $\mathbf{w}_{s}$ and $\mathbf{w}_{o}$ are their parameters.

\noindent \textit{2) Predicate-to-entity Message Propagation:}
As the predicate node set $\mathcal{V}_{p}$ typically includes a considerable amount of false positive predicate proposals, 
we develop a \textit{confidence-aware} adaptive message propagation for entity nodes update to alleviate such noise effect.
Specifically, we first introduce a confidence gating function to control the information flow from an entity's neighbor $\mathbf{r}_{i\rightarrow j}$ as:
\begin{align}
       \gamma_{i\to j} = \mathcal{T}(s^b_{i\to j}), \; \mathcal{T}(x) 
        =\left\{
        \begin{array}{cc}
        0       & x \leq \beta \\
        \alpha  x-\alpha  \beta & \beta<x<1 / \alpha +\beta \\
        1       & x \geq 1 / \alpha +\beta
        \end{array}
        \right. \label{2_stage_gating}
\end{align}
where $\alpha$ and $\beta$ is learnable threshold parameters. 
The gating function $\mathcal{T}(x)$ is designed for achieving a hard control for the predicate proposals with high or low scores (confidently positive or negative), and a soft control for the predicates with intermediate scores.


For each entity node $\mathbf{e}_i$, we divide its neighboring predicates into two sets: $\mathcal{B}_s(i)$ for $\mathbf{e}_i$ as the \textit{subject} and $\mathcal{B}_o(i)$ for $\mathbf{e}_i$ as the \textit{object}.
We update the entity representation $\mathbf{e}_i$ by aggregating its neighbors' messages:
\begin{align}
    \mathbf{e}_i^{(l+1)} = \mathbf{e}_i^{(l)}  +\phi & \left( \frac{1}{|\mathcal{B}_s(i)|}\sum_{k\in\mathcal{B}_{s}(i)} \gamma_{k}  d_{s} \mathbf{W}_e^\intercal\mathbf{r}^{(l)}_{k} \right.\\
     + &\left. \frac{1}{|\mathcal{B}_o(i)|}\sum_{k\in\mathcal{B}_{o}(i)} \gamma_{k}  d_{o}   \mathbf{W}_e^\intercal\mathbf{r}^{(l)}_{k} \right)     
\end{align}
where $\mathbf{W}_e$ is the parameter of a linear transformation.

In each stage, we typically perform $N_i$ iterations of the above two message propagations to capture context in a sufficiently large scope.

%

\subsubsection{Scene Graph Prediction} 
\label{subsec:predictor}

To generate the scene graph of the given image, we introduce two linear classifiers to predict the class of the entities and predicates based on their refined representations. Concretely, for each relationship proposal, our classifier integrates the final representation of predicates proposal from our BGNN, denoted as $\hat{\mathbf{r}}_{i\to j}$, and a class frequency prior~\cite{zellers_neural_2017}, $\hat{\mathbf{p}}_{\mathbf{r}_{i\rightarrow j}}$, for classification:
\begin{align}
    \mathbf{p}_{\mathbf{r}_{i\rightarrow j}} &= \text{softmax}\left(\mathbf{W}_{rel}^\intercal \hat{\mathbf{r}}_{i\rightarrow j} + \text{log}(\hat{\mathbf{p}}_{\mathbf{r}_{i\rightarrow j}}) \right)\in\mathbb{R}^{\mathcal{C}_p}
\end{align}


For each entity, we introduce a learnable weight to fuse the initial visual features $\mathbf{v}_i$ and enhanced features $\hat{\mathbf{e}}_i$ output by our BGNN. The final entity classification is computed as:
\begin{align}
    \mathbf{p}_{\mathbf{e}_i} &= \text{softmax}(\mathbf{W}_{ent}^\intercal ( \rho\hat{\mathbf{e}}_i + (1-\rho)  \mathbf{v}_i )\in\mathbb{R}^{\mathcal{C}_e} 
\end{align}
where $\rho$ is a weight in $[0,1]$, and $\mathbf{W}_{rel}$ and $\mathbf{W}_{ent}$ are the parameters of two classifiers.

\begin{table*}[!ht]
    \begin{center}
        \resizebox{0.9\textwidth}{!}{
            \begin{tabular}{c|l|cc|cc|cc}
                \toprule
                 \multirow{2}{*}{\textbf{B }} & \multirow{2}{*}{\textbf{Models}}& \multicolumn{2}{c|}{\textbf{PredCls}}       & \multicolumn{2}{c|}{\textbf{SGCls}}     & \multicolumn{2}{c}{\textbf{SGGen}} \\
                	\cmidrule{3-8}
                 &   & \textbf{mR@50}~/~\textbf{100} & \textbf{R@50}~/~\textbf{100} & \textbf{mR@50~/~100} & \textbf{R@50}~/~\textbf{100} & \textbf{mR@50}~/~\textbf{100}  & \textbf{R@50}~/~\textbf{100}    \\ \hline
                \multirow{7}{*}{\begin{tabular}[c]{@{}c@{}} \rotatebox{90}{ VGG16 } \end{tabular}}  & 
                	  Motifs\cite{zellers_neural_2017,tang_learning_2018}& 14.0~/~15.3 & 65.2~/~67.1  &  7.7~/~8.2  & 35.8~/~36.5 & 5.7~/~6.6 & 27.2~/~30.3 \\  
                & FREQ \cite{zellers_neural_2017,tang_learning_2018} & 13.0~/~16.0 & 60.6~/~62.2 & 7.2~/~8.5 & 32.3~/~32.9 & 6.1~/~ 7.1 & 26.2~/~30.1  \\              
                & G-RCNN \cite{yang_graph_2018}                      & ~~-~~/~~-~~ & 54.2~/~59.1 &  ~~-~~/~~-~~ & 31.6~/~29.6 & ~~-~~/~~-~~ &  11.4~/~13.7  \\
                &  VCTree\cite{tang_learning_2018}                   & 17.9~/~19.4 & 66.4~/~68.1 &  10.1~/~10.8 &  38.1~/~38.8 & 6.9~/~8.0  & 27.9~/~31.3 \\
                & RelDN\cite{zhang_graphical_2019}                   & ~~-~~/~~-~~ & 68.4~/~68.4 & ~~-~~ /~~-~~ & 36.8~/~36.8 & ~~-~~ /~~-~~ & 28.3~/~32.7  \\
                & KERN \cite{chen_knowledge-embedded_2019}           & 17.7~/~19.2 & 67.6~/~65.8 &  ~9.4~/~10.0 & 36.7~/~37.4  & 6.4~/~7.3  &  29.8~/~27.1    \\
                & GPS-Net\cite{lin_gps-net_2020}                     & ~~-~~ /~22.8& 66.9~/~68.8 & ~~-~~ /~12.6 & 39.2~/~40.1 & ~~-~~ /~ 9.8 & 28.4~/~31.7  \\ 
                & PCPL\cite{yan_pcpl_2020}        & ~35.2~ /~37.8& 50.8~/~52.6 & ~18.6~ /~19.6 & 27.6~/~28.4 & ~9.5~ /~ 11.7 & 14.6~/~18.6  \\ 
                \midrule
                \multirow{11}{*}{\begin{tabular}[c]{@{}c@{}}\rotatebox{90}{ X-101-FPN } \end{tabular}}  
                & RelDN$^\dagger$                                           & 15.8~/~17.2 & 64.8~/~66.7  & 9.3~/~9.6  & 38.1~/~39.3  & 6.0~/~7.3   & 31.4~/~35.9     \\ 
                & Motifs\cite{tang_unbiased_2020}                    & 14.6~/~15.8 & 66.0~/~67.9 & 8.0~/~8.5  & 39.1~/~39.9  & 5.5~/~6.8 &  32.1~/~36.9     \\
                &  Motifs$^{*}$\cite{tang_unbiased_2020}             & 18.5~/~20.0 & 64.6~/~66.7 & 11.1~/~11.8 & 37.9~/~38.8 & 8.2~/~9.7 & 30.5~/~35.4    \\ 
                & VCTree\cite{tang_unbiased_2020}                    & 15.4~/~16.6 & 65.5~/~67.4 & 7.4~/~7.9 & 38.9~/~39.8 & 6.6~/~7.7 &   31.8~/~36.1    \\    
                &  G-RCNN$^\dagger$                                  & 16.4~/~17.2 & 65.4~/~67.2 & 9.0~/~9.5  & 38.5~/~37.0  & 5.8~/~6.6 &  29.7~/~32.8     \\
                &  MSDN$^\dagger$ \cite{li_scene_2017}               & 15.9~/~17.5 & 64.6~/~66.6 & 9.3~/~9.7  & 38.4~/~39.8  & 6.1~/~7.2 &  31.9~/~36.6  \\   
                &  Unbiased\cite{tang_unbiased_2020}                 & 25.4~/~28.7 & 47.2~/~51.6 & 12.2~/~14.0 & 25.4~/~27.9 & 9.3~/~11.1 &  19.4~/~23.2   \\
                & GPS-Net$^\dagger$                                  & 15.2~/~16.6 & 65.2~/~67.1 & 8.5~/~9.1  & 39.2~/~37.8  & 6.7~/~8.6 &   31.1~/~35.9    \\
                &  GPS-Net$^{\dagger *}$                             & 19.2~/~21.4 & 64.4~/~66.7 & 11.7~/~12.5 & 37.5~/~38.6 & 7.4~/~9.5 &  27.8~/~32.1  \\          
                \cmidrule{2-8} 
                &  \textbf{BGNN}          & \textbf{30.4}~/~\textbf{32.9} & 59.2~/~61.3 & \textbf{14.3}~/~\textbf{16.5} & 37.4~/~38.5  & \textbf{10.7}~/~\textbf{12.6} & 31.0~/~35.8   \\ 
            \bottomrule
            \end{tabular}
        }
    \end{center}
\caption{\textbf{The SGG performance of three tasks with graph constraints setting}. $\dagger$ denote results reproduced with the authors' code. $*$ denotes the resampling \cite{gupta_lvis:_2019} is applied for this model.} 
\label{overall_table} 
\vspace{-0.4cm}
\end{table*}

\subsection{Learning with Bi-level Data Sampling}\label{subsec:learning}

We now present our learning strategy for unbiased scene graph generation. We will first develop a bi-level data sampling strategy to balance the data distribution of entities and predicates, and then describe a multitask loss for learning the adaptive BGNN.

\begin{figure}
	\centering
	\includegraphics[width=7.6cm]{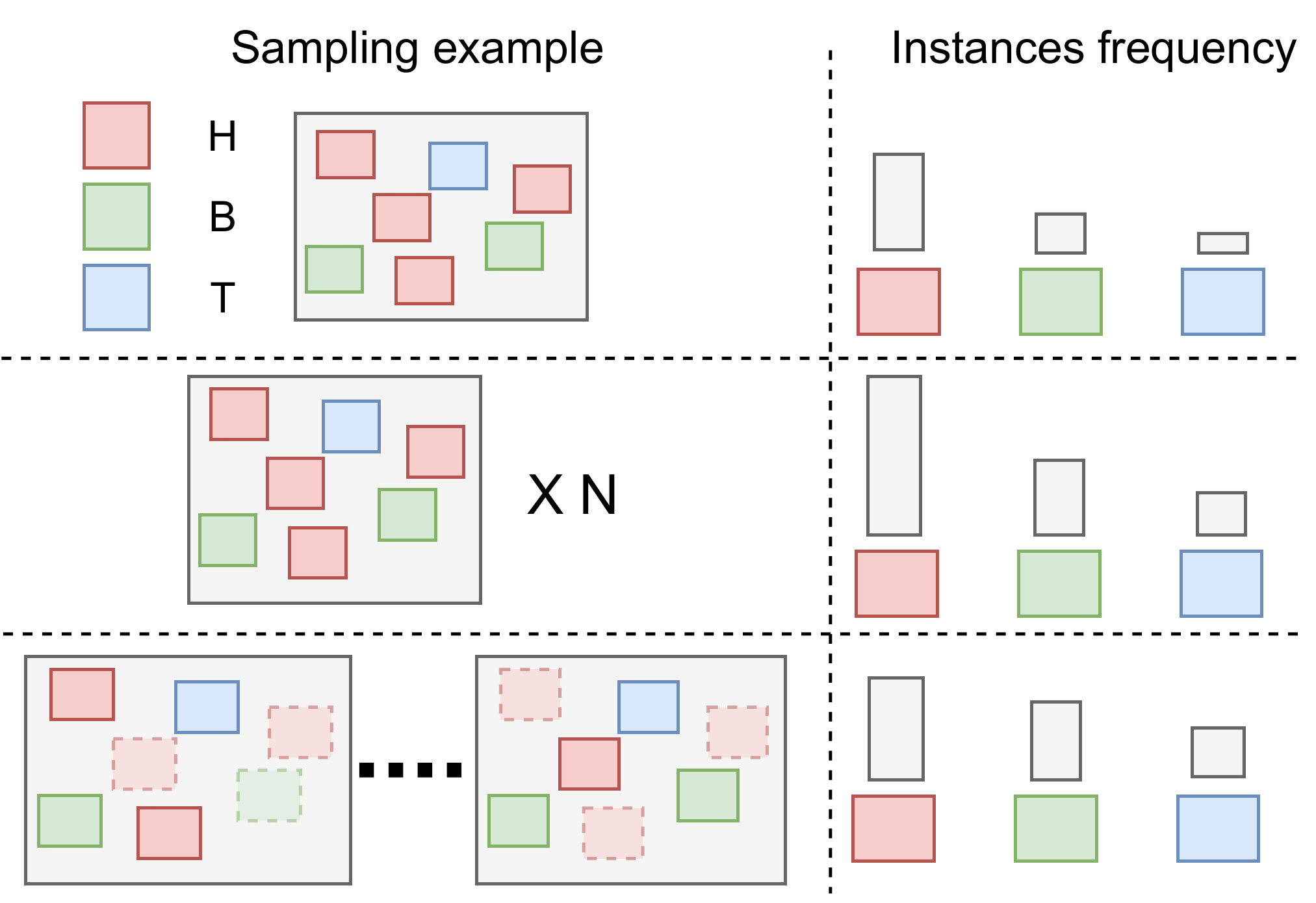}
	\caption{
        \textbf{Illustration of bi-level data sampling for one image.} The top row is the instance frequency of head(H), body(B), and tail(T) categories in the image.
        The middle row shows image-level oversampling with repeat factor $N$. The bottom row shows the instance-level under-sampling for instances of different categories. 
	} 
	\label{data_sampling}
    \vspace{-4mm}
\end{figure}

\vspace{-4mm}
\paragraph{Bi-level Data Resampling}
Unlike in other vision tasks, the scene graph annotations have varying structures, which makes it non-trivial to adopt either the instance-level replay strategy~\cite{hu_learning_2020} or images-level resampling method ~LVIS\cite{gupta_lvis:_2019}.
To tackle the intrinsic long-tail data distribution of entity and relation, we design a two-level data sampling strategy that integrates the above two ideas on rebalancing.  
Specifically, our data sampling strategy consists of two steps: 

\noindent\textit{1) Image-level over-sampling:} We adopt the {repeat factor sampling} in~\cite{gupta_lvis:_2019} to sample images first. We start from a class-specific repeat number, $r^c=\max(1,\sqrt{t/f^c})$, where $c$ is the category, $f^c$ is its frequency on the entire dataset and $t$ is a hyper-parameter that controls when oversampling starts. 
For $i$-th image, we set $r_i = \max_{c \in i} r^c$, where $\{c \in i\}$ are the categories labeled in $i$.

%
%

\noindent\textit{2) Instance-level under-sampling:}  
Given the sampled images, we further design an instance-level sampling strategy for predicates.
Concretely, we compute a drop-out probability for instances of different predicate classes in each image. The drop-out rate $d^c_i$ for instances in $i$-th image, with category label $c$ is calculated by $d^c_i=\max((r_i - r^c) / r_i * \gamma_d, 1.0)$,  and $\gamma_d$ is the hyper-parameter for adjusting the drop-out rate.
With this strategy, our two-level data resampling can achieve an effective trade-off between the head and tail categories.

\vspace{-4mm}
\paragraph{Training Losses}
To train our BGNN model, we design a multitaks loss that consists of three components, including $\mathcal{L}_{rce}$ for relation confidence estimation module (RCE), $\mathcal{L}_{p}$ for predicate proposal classification and $\mathcal{L}_{e}$ for entity proposal classification. Formally, 
\begin{align}
    \mathcal{L}_{total} =\mathcal{L}_{p}  + \lambda_{rce} \mathcal{L}_{rce} +\lambda_{e} \mathcal{L}_{e}
\end{align}
where $\lambda_{rce}, \lambda_{e}$ are weight parameters for calibrating the supervision from each sub-task. 

Here $\mathcal{L}_{p}, \mathcal{L}_{e}$ are the standard cross entropy loss for multi-class classification (foreground categories plus background). The loss of RCE $\mathcal{L}_{rce}$ is composed by two terms:
$\mathcal{L}_{rce} = \mathcal{L}_{m}+\lambda \cdot \mathcal{L}_{b}$,
where $\lambda$ is a weight parameter, and $\mathcal{L}_{m}$ and $\mathcal{L}_{b}$ are losses for the class-specific and overall relation confidence estimation $\mathbf{s}^m,s^b$ respectively. 
Both predictions have explicit supervision as in the training of relationship predictor in the graph refinement stage.
We adopt the focal loss~\cite{lin_focal_2017} to alleviate positive-negative imbalance in the relationship confidence estimation.

\section{Experiments}

In this section, we conduct a series of comprehensive experiments to validate the effectiveness of our method. Below we first present our experimental analysis and ablative study on Visual Genome~\cite{krishna2017visual} dataset (in Sec.~\ref{lab:vg_exp}), then our results on the Open Images V4/V6~\cite{OpenImages} (In Sec.~\ref{lab:oi_exp}). 
In each dataset, we first introduce the implementation details of our method and then report comparisons of quantitative results in detail.






\begin{figure}
    \centering
    \includegraphics[width=7.6cm]{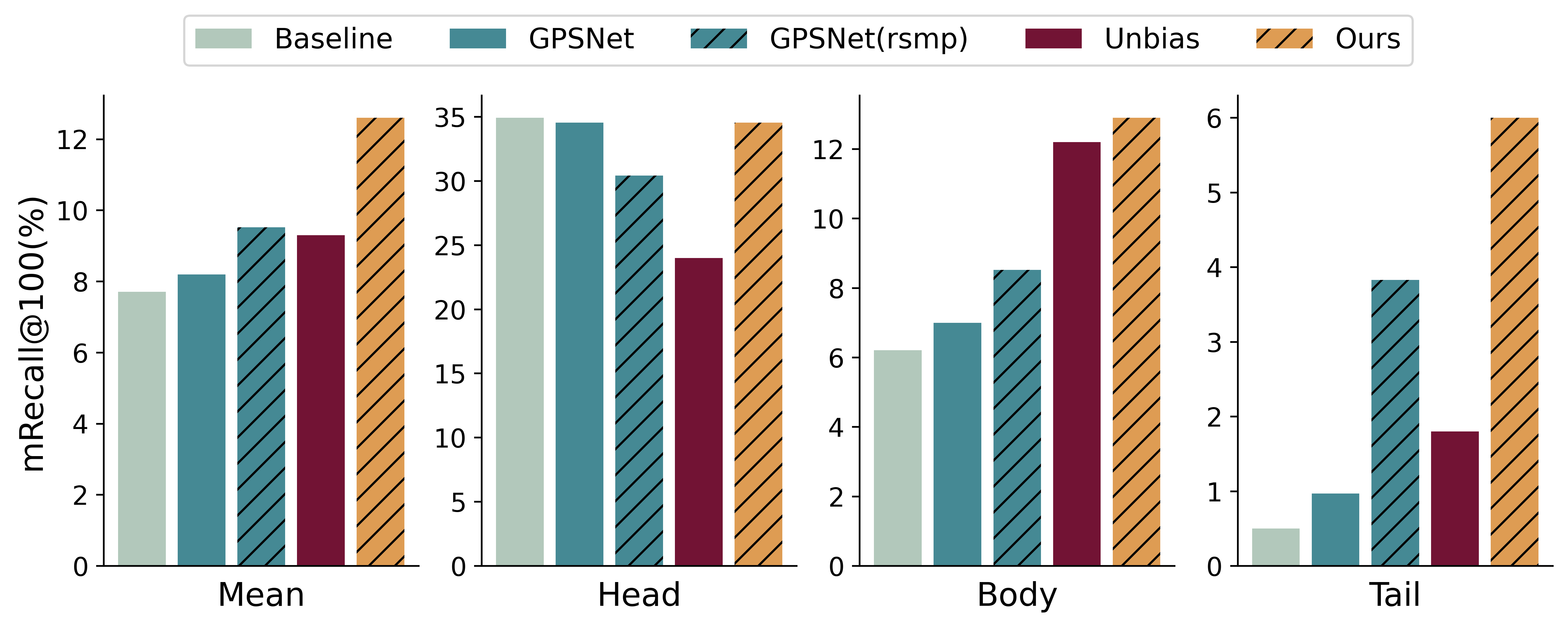}
    \caption{\textbf{The per-group results on VG dataset (SGGen task).}} 
    \label{fig:vg_longtail_part_perf} 
    \vspace{-0.4cm}
\end{figure}

\subsection{Visual Genome} \label{lab:vg_exp}

\subsubsection{Experiments Configurations}

\paragraph{Dataset Details}  
For Visual Genome~\cite{krishna2017visual} dataset, we take the same split protocol as~\cite{xu_scene_2017,zellers_neural_2017}
The most frequent 150 object categories and 50 predicates are adopted for evaluation. Following the similar protocol in~\cite{liu_largescale_2019}, we divided the categories into three disjoint groups according to the instance number in training split: \textit{head}~(more than 10k), \textit{body}~(0.5k $\sim $ 10k), \textit{tail}~(less than 0.5k), more details are reported in Suppl.

\vspace{-0.4cm}

\paragraph{Evaluation Protocol}
We evaluate the model on three sub-tasks as\cite{xu_scene_2017,zellers_neural_2017}:
1) predicate classification (PredCls), 
2) scene graph classification (SGCls),
3) scene graph generation (SGGen, also denote as SGDet).
Following the previous works~\cite{chen_knowledge-embedded_2019, tang_learning_2018,tang_unbiased_2020,lin_gps-net_2020} that concentrate on the long-tail distribution of entity and predicate categories on Visual Genome,
we takes \textbf{recall@K(R@K)} and \textbf{mean recall@K (mR@K)} as evaluation metrics, and also report the mR@K on each long-tail category groups~(\textit{head}, \textit{body} and \textit{tail}).

\vspace{-0.4cm}
\paragraph{Implementation Details}
Similar to Tang~\cite{tang_unbiased_2020}, we adopt ResNeXt-101-FPN~\cite{xie2017aggregated} as backbone network and Faster-RCNN~\cite{ren_faster_2015} as object detector, 
whose model parameters are kept frozen in model training.
In our resampling strategy, we set the repeat factor $t$=0.07, instances drop rate $\gamma_d$=0.7, and weight of fusion the entities features $\rho$=-5.
The $\alpha, \beta$ are initialized as 2.2 and 0.025 respectively.



\subsubsection{Comparisons with State-of-the-Art Methods} \label{subsec:sota_comp}



As shown in Tab.~\ref{overall_table}, BGNN achieves the state of the art in all three sub-tasks (PredCls, SGCls, SGGen) evaluated by mR@50/100 on X-101-FPN backbone, and outperforms Unbiased~\cite{tang_unbiased_2020} with a significant margin of \textbf{5.0} and \textbf{4.2} on PredCls. 
Besides, BGNN shows a comparable performance with previous SOTA in SGCls and SGGen sub-tasks on R@50/100, which demonstrate the effectiveness of our methods.

On the VGG16 backbone, we achieves comparable result with the state-of-the-art, PCPL~\cite{yan_pcpl_2020} on the SGCls and PredCls sub-tasks by mR@50/100.
However, BGNN outperforms the PCPL on SGGen sub-tasks by a large margin, which shows that our proposed BGNN is capable of handling challenging SGGen task with more noisy proposals.
Moreover, as shown in Fig.~\ref{fig:vg_longtail_part_perf}, we compute the mean recall on each long-tail category groups in SGGen sub-task and find BGNN significantly outperforms the prior works~\cite{lin_gps-net_2020, tang_unbiased_2020} on \textit{tail} group, as a result it achieves the highest mean recall over all categories.
We provide more visualization of our results and comparisons in the Suppl.

\begin{table}[]
    \begin{center}
        \resizebox{0.45\textwidth}{!}{
            \begin{tabular}{ccc|cc|ccc}
            \toprule
            \multicolumn{3}{c|}{\textbf{Module}} & \multicolumn{5}{c}{\textbf{SGGen}}                               \\ \midrule
            \textbf{B}&\textbf{C}&\textbf{M}& \textbf{mR@100} & \textbf{R@100}  & \textbf{Head} & \textbf{Body} & \textbf{Tail} \\ \midrule
            \xmark&\xmark&\xmark & 9.7 &34.1 & 32.1& 9.3  & 3.0 \\\midrule
           \cmark &\xmark&\xmark &10.5 &34.8 &  32.4 & 10.7 & 4.5   \\ 
           \cmark &\cmark&\xmark & 11.7&35.3 &  33.6 & 11.4  & 5.2  \\
           \cmark &\cmark&\cmark & \textbf{12.6}&\textbf{35.8} & \textbf{34.0} &  \textbf{12.9} & \textbf{6.0}  \\
           \midrule
          \bottomrule   
        \end{tabular}
        }
    \end{center}
    \caption{\textbf{Ablation study of model structure.} \textbf{B}: bipartite graph neural network with plain message passing; \textbf{C}:confidence-aware message propagation mechanism;\textbf{M}: multi-sage refinement}
    \label{struct_ablation} 
\end{table}

\begin{table}[]
    \begin{center}
    \resizebox{0.44\textwidth}{!}{
        \begin{tabular}{c|cc|ccc}
        \toprule
       \multirow{2}{*}{\textbf{Resample}} & \multicolumn{5}{c}{\textbf{SGGen}}     \\ 
       \cmidrule{2-6}
            &\textbf{mR@100} & \textbf{R@100}  & \textbf{Head} & \textbf{Body} & \textbf{Tail} \\ \midrule
        None 
            & 9.7 &  36.1   &   34.2   &  9.9  & 2.6\\
        RFS\cite{gupta_lvis:_2019} 
            & 10.7 &  34.6  &   33.2   & 9.7  &  3.5 \\ 
        \textbf{BLS} 
            & \textbf{12.6} & \textbf{35.8}   & \textbf{34.0} & \textbf{ 12.9} & \textbf{6.0} \\ 
        \bottomrule
        \end{tabular}
    }
    \caption{
        \textbf{The ablation for the resampling strategy.} RFS: repeat factor sampling; BLS: bi-level data sampling.} 
    \label{tab:resampling}
    \vspace{-0.4cm}
\end{center}
\end{table}

\subsubsection{Ablation Study}
\label{subsec:abla_sty}


\paragraph{Model Components}
We first evaluate the importance of each component in our BGNN. 
As shown in Tab.~\ref{struct_ablation}, we incrementally add one component to the plain baseline to validate their effectiveness.
We observe the plain message propagation(without adaptive confidence-aware mechanism) on bipartite graph neural network improves the baseline with \textbf{8}\% and achieves \textbf{10.5} on mR@100.
Besides, the confidence-aware message propagation mechanism promotes the performance to \textbf{11.7},
and multiple stages refinement further improves the final results to \textbf{12.6}.


\vspace{-4mm}
\paragraph{Re-sampling Strategies}
We compare the widely-used repeat factor sampling~\cite{gupta_lvis:_2019} with BLS for validating the proposed bi-level data sampling strategy~(BLS).
As shown in in Tab.~\ref{tab:resampling}, we find our proposed BLS outperforms the baseline and RFS with a large margin, especially on \textit{body} and \textit{tail} categories.
Besides, unlike other methods, BLS maintains the performance on head categories. This indicates our sampling can balance the prediction on all category groups.

\vspace{-4mm}
\paragraph{Stages and Iterations of BGNN}
To investigate the multiple stage refinement and iterative context encoding in BGNN, we incrementally apply several sets of hyper-parameters on number of stages $N_t$ and iterations $N_i$ in model design.
The quantitative results in Tab.~\ref{tab:iter_num} indicate the message propagation with 3 iterations achieves the best performance in one-stage BGNN.
Furthermore, by freezing the iteration number $N_i$ as 3, we find the performance increases with more stages, and will saturate when $N_t$=3.




\begin{table}[]
    \centering
    \resizebox{7.2cm}{!}{
    \begin{tabular}{cc|cc|ccc}
    \toprule
          &        & \multicolumn{5}{c}{\textbf{SGGen}}                        \\ 
    \midrule
    $N_t$ & $N_i$ &  \textbf{mR@100} & \textbf{R@100} & \textbf{Head} & \textbf{Body} & \textbf{Tail} \\
    \midrule
    1 & 1 &   10.0 & 35.3 &  33.5  &  9.3  &  4.0  \\
    1 & 2 &  10.5 & 35.5 &  34.0  &  9.4    &  4.2  \\
    1 & 3 &  \textbf{10.8} & \textbf{35.3} &  \textbf{33.8}  &  \textbf{10.6}  &  \textbf{4.6}  \\\midrule 
    1 & 3   &  10.8 & 35.3 &  33.8  &  10.6  &  4.6  \\
    2 & 3   &  11.1 & 35.6 &  34.0  &  11.3  &  5.3   \\
    3 & 3   &  \textbf{12.6} & \textbf{35.8} &  \textbf{34.0}  &  \textbf{12.9}  &  \textbf{6.0}    \\ 
    4 & 3   &  12.5 & 35.2 &  34.4  &  12.2  &  5.7    \\
    \bottomrule
\end{tabular}
    }
    \caption{\textbf{The ablation for the different graph iteration parameters.} 
    The ablation of different iteration number for iterative refinement models in our methods. }
    \label{tab:iter_num} 
    \vspace{-0.3cm}

\end{table}

\begin{table}[]
    \centering
    \resizebox{0.47\textwidth}{!}{
        \begin{tabular}{l|l|cc|cc|c}
            \toprule 
            \multirow{2}{*}{\textbf{D }} & \multirow{2}{*}{\textbf{Models}}  &\multirow{2}{*}{\textbf{mR@50}}&  \multirow{2}{*}{\textbf{R@50}} &\multicolumn{2}{c|}{\textbf{wmAP}} & \multirow{2}{*}{\textbf{score}\scriptsize{wtd}}   \\ 
                            \cmidrule{5-6}
                            &  &  &   & \textbf{rel}& \textbf{phr} &   \\ 
                            \midrule
        \multirow{6}{*}{\rotatebox{90}{V4}} 
                            & RelDN\cite{lin_gps-net_2020}  
                                        & -    & 74.94  & 35.54 & 38.52 & 44.61     \\
                            & GPS-Net\cite{lin_gps-net_2020}  
                                        & -    & 77.27  & 38.78 & 40.15 & 47.03     \\  \cmidrule{2-7} 
                            & RelDN $^\dagger$ & 70.40  & 75.66 & 36.13 & 39.91 & 45.21     \\
                            & GPS-Net$^\dagger$  & 69.50  & 74.65  & 35.02 & 39.40 & 44.70     \\
                            \cmidrule{2-7} 
                            & \textbf{BGNN} & \textbf{72.11} & 75.46 & 37.76 & 41.70 & 46.87    \\ 
                            \midrule
        \multirow{10}{*}{\rotatebox{90}{V6}} 
                            & RelDN $^\dagger$  & 33.98 & 73.08 & 32.16 & 33.39 & 40.84   \\
                            & RelDN $^{\dagger*}$ & 37.20 & 75.34 & 33.21 & 34.31 & 41.97 \\
                            & VCTree$^\dagger$  &  33.91 & 74.08 & 34.16 & 33.11 & 40.21   \\
                            & G-RCNN$^\dagger$  &34.04& 74.51 & 33.15 & 34.21 & 41.84   \\
                            & Motifs$^\dagger$ & 32.68 &71.63& 29.91 &31.59   & 38.93    \\
                            & Unbiased$^\dagger$ & 35.47 & 69.30 & 30.74  & 32.80  & 39.27   \\ 
                            & GPS-Net$^\dagger$ & 35.26 & 74.81 & 32.85&33.98 & 41.69 \\
                            & GPS-Net$^{\dagger*}$ & 38.93 &74.74 & 32.77  &33.87 & 41.60    \\ \cmidrule{2-7} 
                            & \textbf{BGNN}    & \textbf{40.45} & 74.98  & 33.51    & 34.15   & \textbf{42.06}     \\ 
                            \bottomrule
        \end{tabular}
    }
    \caption{\textbf{The Performance table of Open Images Dataset.} 
    $*$ denotes the resampling~\cite{gupta_lvis:_2019} is applied for this model.
    $\dagger$ denote results reproduced with the authors' code. } 
    \label{tab:openimage}
    \vspace{-0.4cm}
\end{table}

\subsection{Open Images} \label{lab:oi_exp}

\subsubsection{Experiments Setting}

\paragraph{Dataset Details.} 

The \textbf{Open Images} dataset~\cite{OpenImages} is a large-scale dataset proposed by Google recently. Compared with Visual Genome dataset, it has a superior annotation quality for the scene graph generation. In this work, we conduct experiments on Open Images  V4\&V6, we follow the similar data processing and evaluation protocols in~\cite{zhang_graphical_2019, OpenImages, lin_gps-net_2020}. 

The Open Images V4 is introduced as a benchmark for scene graph generation by~\cite{zhang_graphical_2019} and~\cite{lin_gps-net_2020}, which has 53,953 and 3,234 images for the train and val sets, 57 objects categories, and 9 predicate categories in total.
The Open Images V6 has 126,368 images used for training, 1813 and 5322 images for validation and test, respectively, with 301 object categories and 31 predicate categories. 
This dataset has a comparable amount of semantics categories with the VG.

\vspace{-4mm}
\paragraph{Evaluation Protocol}\label{subsec:eval_prot}
For Open Images  V4\&V6, we follow the same data processing and evaluation protocols in~\cite{zhang_graphical_2019, OpenImages, lin_gps-net_2020}. 
The mean Recall@50 (mR@50), Recall@50 (R@50), weighted mean AP of relationships ( $\text{wmAP}_{rel}$), and weighted mean AP of phrase ($\text{wmAP}_{phr}$) are used as evaluation metrics. 
Following standard evaluation metrics of Open Images refer to~\cite{OpenImages, zhang_graphical_2019, lin_gps-net_2020}, the weight metric $\text{score}_{wtd}$ is computed as: $\text{score}_{wtd} = 0.2 \times \text{R}@50 +0.4 \times \text{wmAP}_{rel} + 0.4 \times \text{wmAP}_{phr}$.
Besides, we also report mRecall@K like Visual Genome as a balanced metric for comprehensive comparison.



\subsubsection{Quantitative Results}

Performance on these two datasets are reported in Tab.~\ref{tab:openimage}. 
For Open Images V4, which includes only 9 predicate categories in total, make it constrained to explore the biased scene graph generation tasks. 
Our method can still achieve competitive results on weighted metric $\text{score}$, and outperform the previous work on mean recall with a significant margin.
For Open Images V6 with 31 predicate categories,
we reimplement several recent works~\cite{lin_gps-net_2020,tang_unbiased_2020, tang_learning_2018, zellers_neural_2017, yang_graph_2018} for fair comparison. As shown in Tab.~\ref{tab:openimage}, our method achieves the state-of-the-art performance on mean recall and competitive results on weighted metric $\text{score}$. Results on both datasets demonstrate the efficacy of our approach.

\section{Conclusion}
In this work, we have proposed a novel bipartite graph neural network (BGNN) for unbiased scene graph generation. Compared to previous methods, our main contribution consists of two key components as follows. We first develop a confidence-aware message passing mechanism for our BGNN to encode scene context in an effective manner; Moreover, we design a bi-level resampling strategy to mitigate the imbalanced data distribution during the training.
The results evidently show that our BGNN achieves the superior or comparable performances over the prior state-of-the-art approaches on all three scene graph datasets.


{\small
\bibliographystyle{ieee_fullname}
\bibliography{egbib}
}

\newpage

\appendix

\section{Relation Confidence Estimation Module}
As introduced in the main paper, the RCE module is an important parts of our method.
To demonstrate the effectiveness of the RCE module, we further introduce the learning details of the RCE module and performance comparison with the similar model proposed by previous works.

\subsection{Learning}
We use a supervised learning strategy to train the RCE module of BGNN, in which the predicate class labels (which predicate category and whether it is valid predicate or background) are used for supervision. 
Different from the cross-entropy loss $\mathcal{L}_p$ used for the final predicate predictions, we develop a multi-task loss $\mathcal{L}_{rce}$ for the RCE module. Specifically, we use two confidence predictions from the RCE: multi-categories confidence score $\mathbf{s}^m \in \mathbb{R}^{C_p}$ and binary confidence score $s^b$. We define two focal losses, $\mathcal{L}_{m},  \mathcal{L}_{b}$ on the confidence predictions of $M$ predicate proposals $\{\mathbf{s}^m_1, ... \mathbf{s}^m_M\}, \{s^b_1, ... s^b_M\}$, respectively. 
Formally:
\begin{align}
\mathcal{L}_{m} &= -\alpha\frac{1}{M}\sum_k^{M}\sum_i^{|C_{p}|} \mathbf{y}_{k, i} (1- \mathbf{s}^m_{k,i})^{\gamma} \cdot \log(\mathbf{s}^m_{k,i}) \\
\mathcal{L}_{b} &= -\alpha\frac{1}{M}\sum_k^{M} y'_k(1- s^b_{k})^{\gamma} \cdot \log(s^b_{k})
\end{align}

where $\mathbf{y}_{k}$ is one-hot vector and $y'_k$ is binary label of positive predicate proposals. $\alpha,\gamma$ are the hyper-parameters.

\subsection{Performance}
To demonstrate the effectiveness of the RCE module on removing negative predicate proposals, we use the \textbf{AUC} to measure its performance, and compare it with two alternatives: \textit{production of entities prediction score} and \textit{relation proposal network} proposed by Graph-RCNN.
The AUC of those three methods are \textbf{0.839, 0.629, 0.671}, respectively on the validation set of VG, which indicates RCE module is more effective than previous works.

\begin{figure}
   \centering
   \includegraphics[width=0.95\linewidth]{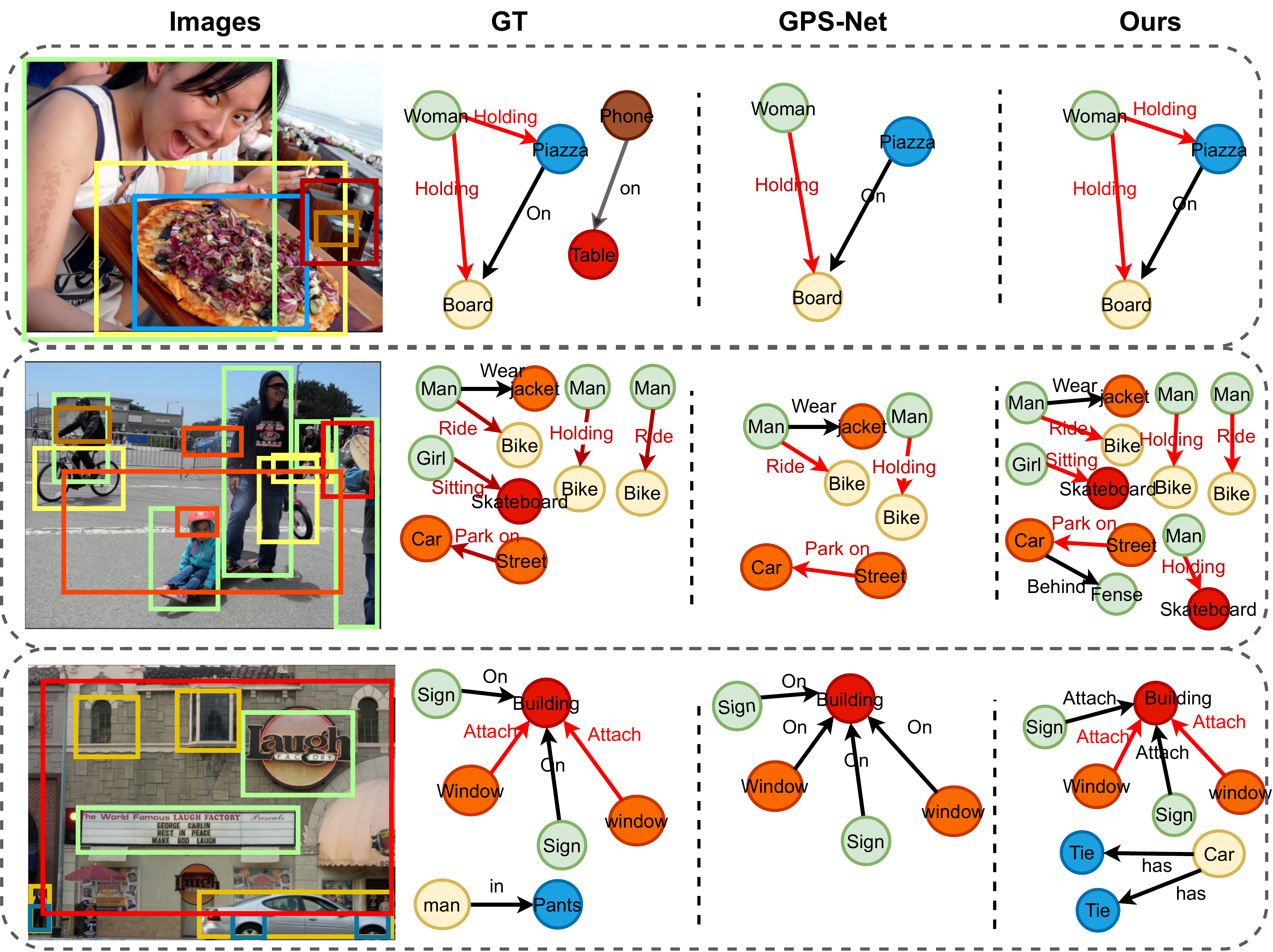}
   \caption{\textbf{Qualitative comparisons between our method and GPS-Net$\dagger$ in the SGGen setting.}
   The predicates in \textit{body} and \textit{tail} categories group are marked as \textcolor{red}{red} color.
   We also show the reasonable relationships detected by models which are not included in GT.}  
   \label{fig:qualitative}  
\end{figure}

\section{Model Comparison w/ Resampling}
We note that we have reported the SOTA with recent RFS resampling in Tab.1. To further demonstrate the effectiveness of our BGNN, we add our BLS to other recent methods (reimplemented GPS-Net and MSDN) and perform comparisons on the SGGen task as below:
\begin{table}[ht]
   \centering
       \resizebox{0.40\textwidth}{!}{

         \begin{tabular}{l|cc|ccc}
            \toprule
            \multirow{2}{*}{\textbf{Models}} & \multicolumn{5}{c}{\textbf{SGGen}}     \\ 
            \cmidrule{2-6}
                 &\textbf{mR@100} & \textbf{R@100}  & \textbf{Head} & \textbf{Body} & \textbf{Tail} \\ \midrule
            GPS-Net $_{\text{w/ BLS}}$ & 11.4 & 34.3   & 32.3 & 9.9  & 4.0           \\
            MSDN $_{\text{w/ BLS}}$  & 11.8 & 34.4    & 32.4 & 10.5 & 5.1           \\
            \midrule 
            BGNN $_{\text{w/ BLS}}$  & \textbf{12.6} & \textbf{35.8}  & \textbf{34.0}  & \textbf{12.9} & \textbf{6.0}    \\
            \bottomrule    
         \end{tabular}
      }\caption{\textbf{The performance comparison between SOTA with our BLS.}}
\end{table}

The results show that our BGNN still outperforms other approaches under the same resampling strategy. In addition, we emphasize that the bi-level resampling is also our main contribution, and the above results demonstrate its effectiveness for improving all three methods.

\section{Quantitative Studies}
We extend the quantitative studies as a supplement to the main paper.
In this section, we show the detail of long-tail parts partition, and performance comparison on each long-tail part in Sec~\ref{lcgp}.
For the fair comparison with the previous methods, we also show the per-class performance comparison on the PredCls subtask in Sec~\ref{ppc}.
In Sec~\ref{visualize}, we show the comparison of model prediction by visualizing the scene graph generated by BGNN and previous SOTA GPS-Net.

\subsection{Long-tail Categories Groups Partition} \label{lcgp}

\begin{figure*}[hb]
	\centering
	\includegraphics[width=\linewidth]{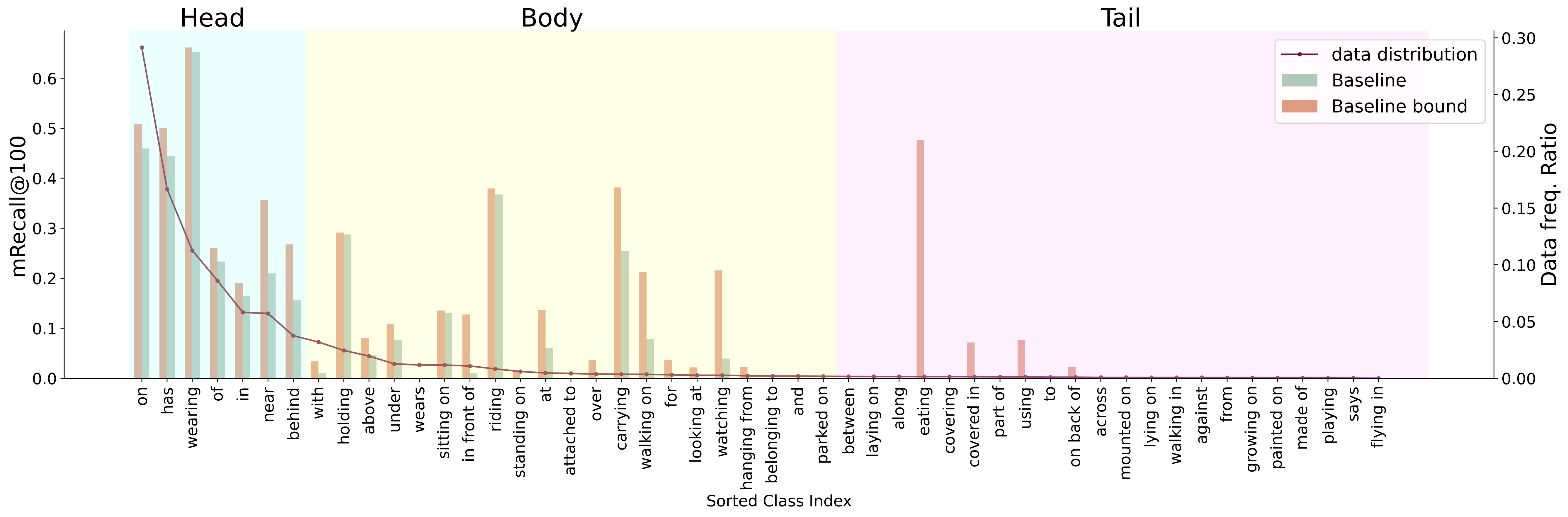}
	\caption{
		 \textbf{The long-tail categories groups partition and the upper-bound comparison on Visual Genome dataset.} 
	}\label{fig:vg_longtail_part}
\end{figure*}

\paragraph{Visual Genome}
First, we report the data distribution and long-tail categories set partition detail of Visual Genome~\cite{krishna2017visual, xu_scene_2017} in figure \ref{fig:vg_longtail_part}. 
We divide the categories into three disjoint groups according to the instance number in training split: \textit{head}(more than 10k), \textit{body}(0.5k $\sim $ 10k), \textit{tail}(less than 0.5k)

We further present the performance comparison of the baseline model(MSDN) between our upper bound assumption referred to Sec. 1 of main paper. 
The result indicates reducing noise in context modeling improve the baseline model with a large margin especially on tail categories, which only have several data points.

\paragraph{Open Images}

\begin{figure*}
	\centering
	\includegraphics[width=\linewidth]{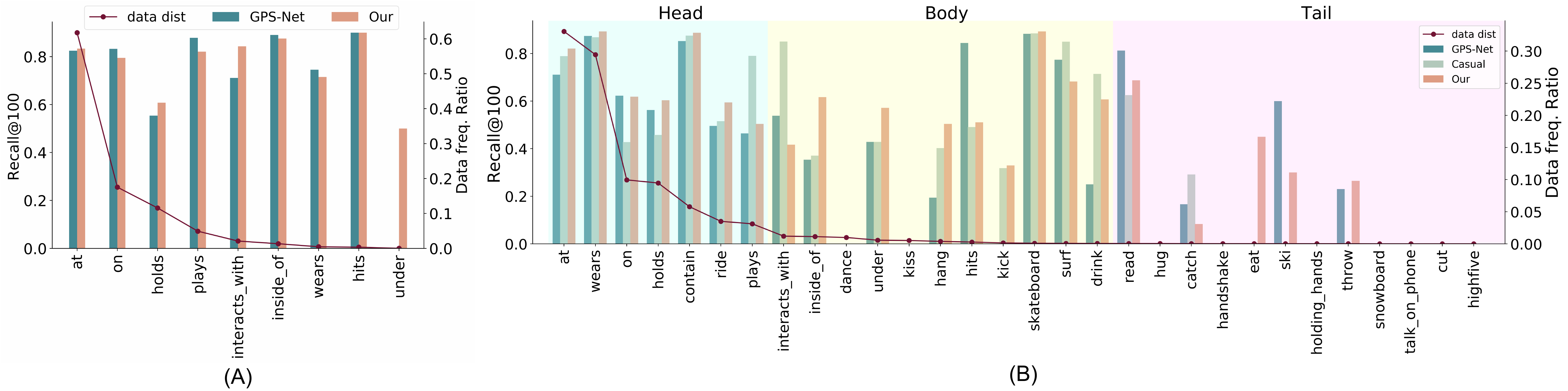}
	\caption{
		\textbf{The long-tail categories groups partition and per-class performance comparison of Open Images dataset}. Part (A) is the Open Images V4, part (A) is the Open Images V6 dataset. We compare with the two SOTA methods: Causal~\cite{tang_unbiased_2020}, and GPS-Net~\cite{lin_gps-net_2020}.
	}
	\label{fig:oi_dataset_alt}
\end{figure*}

The long-tail categories group partition and per-class performance comparison on Open Images dataset are reported in Fig. \ref{fig:oi_dataset_alt}. 
Similarly, we divide the categories of Open Images V6 into three groups according to the instance number in training split: \textit{head}(more than 12k), \textit{body}(0.2k $\sim $ 12k), \textit{tail}(less than 0.2k). 
For performance comparison with the SOTA method, our method achieves significant improvement on tail categories and achieves the comparable overall performance with the GPS-Net~\cite{lin_gps-net_2020} and Causal~\cite{tang_unbiased_2020}.

\begin{figure*}
	\centering
	\includegraphics[width=\linewidth]{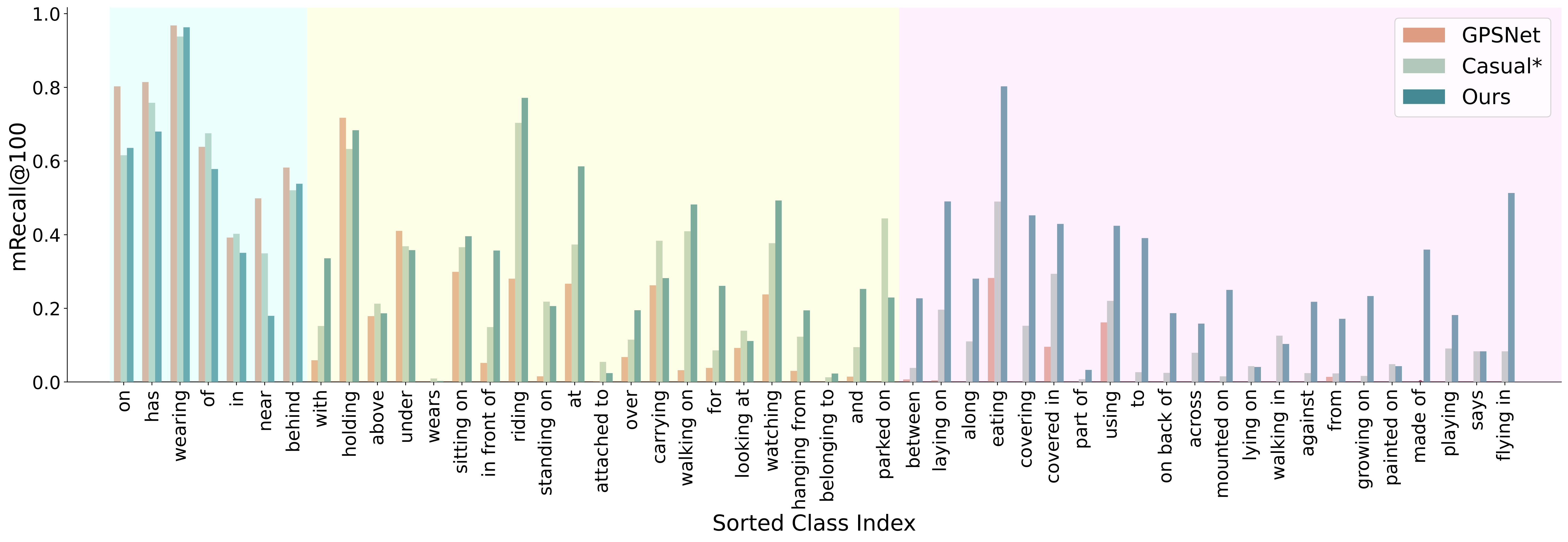}
	\caption{
		 \textbf{The Recall@100 on Predicate Classification(PredCls) of all categories.} We compare with the SOTA methods: Causal~\cite{tang_unbiased_2020}, and GPS-Net~\cite{lin_gps-net_2020}.
		 $*$ denotes the re-sampling~\cite{gupta_lvis:_2019} is applied for this model.
	}
	\label{fig:predcls_cmp}
\end{figure*}

\subsection{Per-class Performance Comparison with the Other Models} \label{ppc}

Following the previous works setting~\cite{chen_knowledge-embedded_2019,tang_learning_2018, lin_gps-net_2020, tang_unbiased_2020}, we show the comparison of Recall@100 on PredCls sub-task of each categories with the two SOTA methods~\cite{lin_gps-net_2020, tang_unbiased_2020}, as shown in fig \ref{fig:predcls_cmp}.

Instead of only comparing the top-35 frequency categories, we present all 50 categories of Visual Genome. Our model achieves a significant performance gain on low-frequency categories, which demonstrates the effectiveness of our BGNN.

\subsection{Visualization of Model Prediction} \label{visualize}
To better understand the BGNN, we visualize scene graph generation prediction from the Visual Genome dataset.
As shown in Fig.~\ref{fig:qualitative}, our model has a significant improvement for \textit{body} and \textit{tail} categories group compared with GPS-Net.
With a more effective confidence-aware message propagation mechanism, our model has better context modeling capability of visual representations for low-frequency categories.

\end{document}